\documentclass[journal]{vgtc}                     

\onlineid{0}

\vgtccategory{Research}

\vgtcpapertype{please specify}

\title{SPAC-Net: Rethinking Point Cloud Completion with Structural Prior}

\author{%
  \authororcid{Zizhao Wu}{0000-0002-1825-0097},
  Jian Shi, Xuan Deng, Cheng Zhang, Genfu Yang, 
  Ming Zeng and Yunhai Wang
}

\authorfooter{
  \item
  	Zizhao Wu, Jian Shi, Xuan Deng, Cheng Zhang and Genfu Yang are with the School of Digital Media Technology, Hangzhou Dianzi University, Hangzhou 310018, China (Email: wuzizhao@hdu.edu.cn; sand2sand@hdu.edu.cn; dengxuan08@hdu.edu.cn; zhangcheng828@hdu.edu.cn; ygfsn@hdu.edu.cn) (Corresponding author: Genfu Yang.).
  \item
  	Ming Zeng is with the School of Informatics, Xiamen University, Xiamen 361000, China
  	E-mail: zengming@xmu.edu.cn
   \item
  	Yunhai Wang is with the School of Information, Renmin University of China, Beijing, 100086, China
  	E-mail: cloudseawang@gmail.com
    \item   Our implementation is publicly available at: https://github.com/sand2sand/SPAC-Net.
}

\abstract{%
 Point cloud completion aims to infer a complete shape from its partial observation. Many approaches utilize a pure encoder-decoder paradigm in which complete shape can be directly predicted by shape priors learned from partial scans, however, these methods suffer from the loss of details inevitably due to the feature abstraction issues. In this paper, we propose a novel framework, termed SPAC-Net, that aims to rethink the completion task under the guidance of a new structural prior, we call it \emph{interface}. Specifically, our method first investigates Marginal Detector (MAD) module to localize the \textit{interface}, defined as the intersection between the known observation and the missing parts. Based on the \textit{interface}, our method predicts the coarse shape by learning the displacement from the points in \textit{interface} move to their corresponding position in missing parts. Furthermore, we devise an additional Structure Supplement (SSP) module before the upsampling stage to enhance the structural details of the coarse shape, enabling the upsampling module to focus more on the upsampling task. Extensive experiments have been conducted on several challenging benchmarks, and the results demonstrate that our method outperforms existing state-of-the-art approaches.
}


\graphicspath{{figs/}{figures/}{pictures/}{images/}{./}} 

\usepackage{tabu}                      
\usepackage{times}  
\usepackage{helvet}  
\usepackage{courier}  
\usepackage[hyphens]{url}  
\usepackage{graphicx} 
\usepackage{caption} 
\usepackage{booktabs}
\usepackage{algorithm}
\usepackage{algorithmic}
\usepackage{amsmath}
\usepackage{array}
\usepackage{multirow}
\usepackage[switch]{lineno}
\makeatletter
\newcommand{\thickhline}{%
 \noalign {\ifnum 0=`}\fi \hrule height 0.8pt
\futurelet \reserved@a \@xhline
}
\newcommand{\tabincell}[2]{\begin{tabular}{@{}#1@{}}#2\end{tabular}}
\makeatother
\usepackage{newfloat}
\usepackage{listings}
\usepackage{mathptmx}                  

\begin{document}

\firstsection{Introduction}
\label{Introduction}
\maketitle

With the rapid development of 3D scanning technology such as laser scanners \cite{KITTI} and depth
cameras \cite{DaiCSHFN17}, an increasing amount of point cloud data has emerged. These point cloud data are commonly used as representations of three-dimensional objects and have a wide range of applications including robotics \cite{VarleyDRRA17}, augmented reality \cite{AlexiouUE17}, and automated driving \cite{ZhengLYL22}.
However, due to factors such as occlusion, noise, and limited views of 3D sensors, the obtained point cloud data is usually incomplete and sparse. Therefore, point cloud completion is studied to predict the complete shapes of objects from partial observations.

\begin{figure}[!t]
  \centering
  \setlength{\belowcaptionskip}{-4mm}
  \includegraphics[width = 1\linewidth]{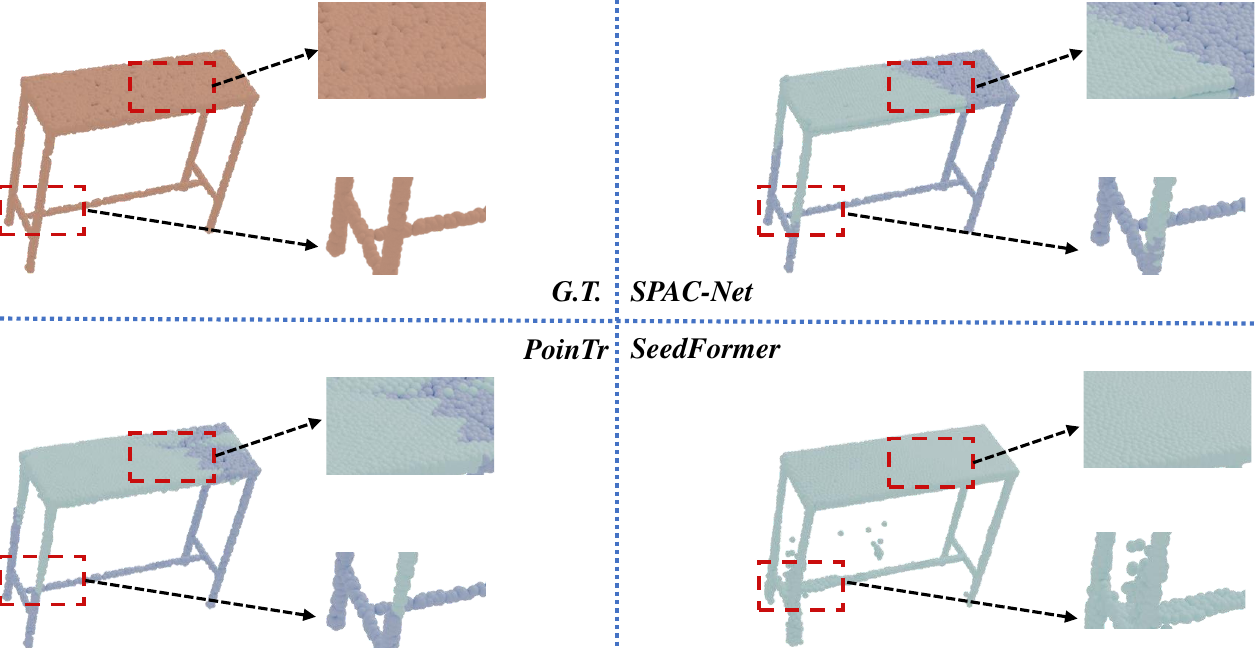}
  \caption{Our method is presented in a comparative visualization with other state-of-the-art (SOTA) point cloud completion techniques. In the comparative figures, two detail areas from the completion targets are selected to illustrate the effectiveness of the different approaches. The cyan points denote the predicted point cloud, whereas the blue points represent the input partial scans.}
  \label{fig:page1_compare_vis}
\end{figure}

Some earlier approaches \cite{AbhishekSharma2016VConvDAEDV, ZhijianLiu2019PointVoxelCF, AngelaDai2016ShapeCU, XiaoguangHan2017HighResolutionSC, StutzDavid2018Learning3S, ChengZLWL19, VarleyDRRA17, DucThanhNguyen2016AFM} have attempted to tackle the task by drawing analogies with 2D completion tasks using voxel and 3D convolution, but these methods have limitations in terms of computational resources. As pioneers in directly processing point cloud data, PointNet \cite{CharlesRQi2016PointNetDL} and its variants \cite{CharlesRQi2017PointNetDH, YueWang2018DynamicGC, PointNetLK, PointNet_3, PointNetXt, LiWF23, ZhangCWYW22} take raw point cloud as input, effectively reducing memory costs. However, they struggle to recover details in their predictions due to the max-pooling operation in feature learning. Recently, witness the power of Transformer architecture, many works \cite{XuminYu2021PoinTrDP, HaoranZhou2022SeedFormerPS} have successfully applied the Transformer \cite{AshishVaswani2017AttentionIA} to point cloud completion. 
Typically, these methods incorporate Transformers into an encoder-decoder architecture, attempting to directly learn the completed result from the input point cloud. However, they often lack sensitivity to the missing parts due to the absence of geometric priors. Additionally, simultaneous upsampling and detail enhancement in many approaches, such as PoinTr \cite{XuminYu2021PoinTrDP} and SeedFormer \cite{HaoranZhou2022SeedFormerPS}, can introduce noise during the generation process. An example is illustrated in the bottom row of Figure \ref{fig:page1_compare_vis}. The completion result obtained by PoinTr on the left shows an uneven density distribution at the boundaries, while the result from SeedFormer on the right exhibits noise in the table leg area.

\begin{figure}[!t]
  \centering
  \setlength{\belowcaptionskip}{-4mm}
  \includegraphics[width = 1\linewidth]{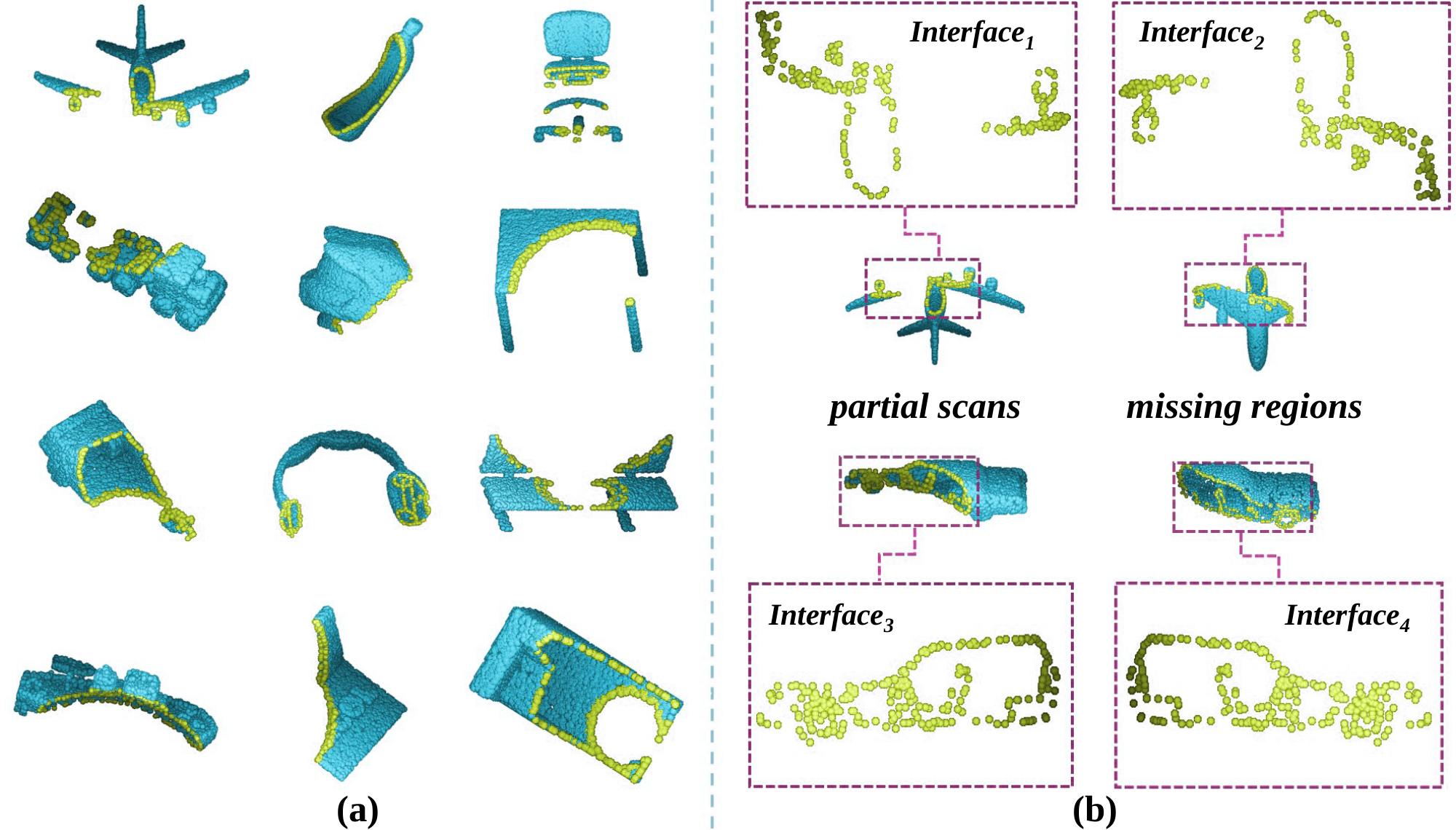}
  \caption{\footnotesize\textbf{(a)} shows several visualization results of \textit{interface}(yellow points) in incomplete object. $\textbf{(b)}$ 
 illustrates the \textit{interface} in partial scans is equivalent to the one in missing parts, which means that $\textit{interface}_{1}$ equals to $\textit{interface}_{2}$, $\textit{interface}_{3}$ equals to $\textit{interface}_{4}$. So we can localize \textit{interface} in partial scans to establish spatial perception on missing parts before predicting them.}
  \label{fig:bd}
\end{figure}

Our primary objective is to enhance the sensitivity of our model to the missing parts of the given partial scans. By being guided by the explicitly missing information, our method aims to perform a more accurate and focused completion of the missing regions without being affected by other noisy information.
To achieve this, we introduce a novel structural prior called \textit{interface} which is defined as the intersection between partial scans and missing parts, as depicted in Figure~ \ref{fig:bd}. In the completed shape, which is a combination of partial scans and missing parts, we observe that they exhibit complementary characteristics in terms of shape, as well as their \textit{interface}. when the radius of each point approaches infinitesimal, we consider the \textit{interface} composed of points from the two regions as equivalent. This implies that obtaining the \textit{interface} of partial scans is equivalent to acquiring the ones of the missing parts. Therefore, we consider the \textit{interface} as a novel structural prior that serves as a guide for our model to perceive the missing parts before predicting them. 
Meanwhile, by learning the displacement of points in the \textit{interface} to their corresponding positions in the missing parts, our method can effectively mitigate the loss of local features caused by the use of max-pooling in many existing methods.

Moreover, many methods \cite{ZitianHuang2020PFNetPF, XuminYu2021PoinTrDP, HaoranZhou2022SeedFormerPS, ProxyFormer} have integrated upsampling modules into their frameworks for the completion task. However, these approaches perform the upsampling operations based on the coarse shape that lacks structural details. Thus, enhancing the structural details of the coarse shape before upsampling is necessary. To solve this, we introduce an additional SSP module, which dynamically refines and optimizes the structural details of the coarse shape, supporting the upsampling module with high-quality input. Consequently, this module facilitates the further restoration of shape details, enhancing the overall performance of the completion process. In Figure 1, we illustrate the results of our proposed method. Compared to other methods, our results better preserve structural characteristics while maintaining a more uniform distribution.

Building upon the aforementioned blocks, we propose a novel encoder-decoder framework called Structural Prior-Guidance Point Cloud Completion Network (SPAC-Net), which divides the completion task into two major phases: shape generation and shape upsampling. In the first stage, after localizing the \textit{interface} through MAD module, our method predicts the coarse shape by learning the displacement that points in \textit{interface} move to their corresponding position in missing parts, thus effectively alleviating the loss of local feature caused by using max-pooling on shape priors. Furthermore, unlike previous methods that directly turn into the upsampling stage after obtaining coarse shape, we introduce an additional SSP module to enhance the structural details before shape upsampling. In the subsequent stage, to encourage uniform dispersion of dense points on the shape surface, we employ a lightweight yet efficient FoldingNet\cite{YaoqingYang2017FoldingNetPC}. 
We conducted extensive experiments on publicly available point cloud completion datasets \cite{XuminYu2021PoinTrDP,WentaoYuan2018PCNPC,KITTI}, and the experimental results demonstrate that our method surpasses previous approaches.

In summary, the primary contributions of our work can be summarized as follows:
\begin{itemize}

    \item We propose a novel framework namely SPAC-Net for the point cloud completion task. Instead of a purely encoder-decoder framework the existing methods commonly adopted, our method investigates a novel structural prior called \textit{interface} to facilitate the task which brings about high quality completion results.
    \item Our method investigates to learn the displacement from the points in the \textit{Interface} move to their corresponding positions in the missing parts, rather than directly generating coarse shapes based on shape priors. We also propose two essential modules MAD and SSP. The MAD module is in charge of localizing the \textit{interface}, while the SSP module is specifically responsible for refining the details of coarse shape. 
    \item The proposed method achieves significant performance compared with the state-of-the-art approaches on challenging benchmarks. 
\end{itemize}

\section{Related Work}

\subsection{Point Cloud Completion }
With the rise of Convolutional Neural Networks (CNNs), many earlier approaches  \cite{AbhishekSharma2016VConvDAEDV, ZhijianLiu2019PointVoxelCF, AngelaDai2016ShapeCU, XiaoguangHan2017HighResolutionSC, StutzDavid2018Learning3S, VarleyDRRA17, DucThanhNguyen2016AFM} have employed 3D convolution for 3D shape completion that uses the voxel grids to represent intermediate 3D shape. Therefore, it is simple to extend powerful 2D CNN baselines to the point cloud analysis tasks. For example, 3D-EPN \cite{AngelaDai2016ShapeCU} presents an encoder-decoder framework that divides each model into 3D voxel grids and applies 3D convolution to perform shape completion task in a multi-resolution synthesis step. Alternatively, GRNet \cite{YingLi2020GRNetGR} maps the points within each voxel to its eight vertices for exploring the structural context, which helps to balance the voxel resolution partly. However, this kind of method still suffers from heavy computational cost and memory consumption, due to the resolution issue of the voxel grids.
Recently, PointNet \cite{CharlesRQi2016PointNetDL} pioneers the point cloud analysis paradigm that directly consumes the 3D coordinates for downstream tasks. 
The framework, along with its extended version including PointNet++\cite{CharlesRQi2017PointNetDH}, has emerged as the prominent approaches in the field.

Following these methods, PCN \cite{WentaoYuan2018PCNPC} first employs lightweight PointNet layers to extract features and then leverages FoldingNet \cite{YaoqingYang2017FoldingNetPC} and fully connected layers to decode the missing parts. PF-Net \cite{ZitianHuang2020PFNetPF} introduces a Multi-Resolutional Encoder (MRE) based on PointNet++ \cite{CharlesRQi2017PointNetDH} to extract features at different resolutions and then concatenated to capture local features. However, the utilization of max-pooling operation on local features in these methods leads to the loss of fine-grained details.
TopNet \cite{LynePTchapmi2019TopNetSP} focuses on improving the decoder by adopting a rooted tree architecture to implicitly model and generate point cloud. Additionally, Wang et al. \cite{YidaWang2022LearningLD} propose a novel pooling mechanism called neighbor-pooling. This mechanism aims to downsample the data in the encoder while preserving individual feature descriptors obtained from the Iterative Closest Point (ICP) algorithm \cite{PaulJBesl1992MethodFR}, which are considered vital for shape completion.
Furthermore, benefiting from the significant performance of Transformer \cite{AshishVaswani2017AttentionIA} backbone, Guo et al. \cite{Guo_Cai_Liu_Mu_Martin_Hu_2021} propose PCT, which successfully applies the Transformer from natural language processing to point cloud processing and achieves impressive performance across various downstream tasks. 
Afterwards, transformer-based works became popular, 
PoinTr\cite{XuminYu2021PoinTrDP} adopts attention mechanisms to prioritize the geometric shape and intricate details of the missing regions during the completion process. 
SeedFormer \cite{HaoranZhou2022SeedFormerPS} firstly utilizes patch seeds to preserve local feature during the generation of coarse shape, then an upsampling transformer is devised as a point generator to complete the completion results, which helps to promote the recovery quality of geometric details by learning the distribution of existing points. 
Zhang et al. \cite{ZhangZDLYX23} propose a two-stage coarse-to-fine completion framework based on skeleton-detail Transformer, which mines the correlation between local patterns and global shape, enhancing the geometric details of the completion results. 
ProxyFormer \cite{ProxyFormer} presents a missing part-sensitive transformer and an alignment strategy to perceive the position of missing parts.
AnchorFormer \cite{AnchorFormer} predicts the intermediate representation anchor points and uses these predictions to determine the position offsets from them to the missing points in the target region, which facilitates preserving local features. The generated results are further refined using the designed morphing block.
P2C \cite{P2C} predicts the occluded blocks by learning shape priors information from different local blocks in a self-supervised manner. Additionally, P2C introduces the Region-Aware Chamfer Distance to regulate shape mismatches without compromising completion capability and incorporates the Normal Consistency Constraint to incorporate the assumption of local planarity, encouraging the recovered shape surface to be continuous and complete.

In this work, to guide the network to focus on the missing areas, we propose a novel shape completion method based on structural prior. Compared to the aforementioned methods, our method focuses more on the completion of missing parts, thereby performing better in preserving sensitive details.

\begin{figure*}[!t]
    \centering
    \setlength{\belowcaptionskip}{-2mm}
    \includegraphics[width = 1.0\linewidth]{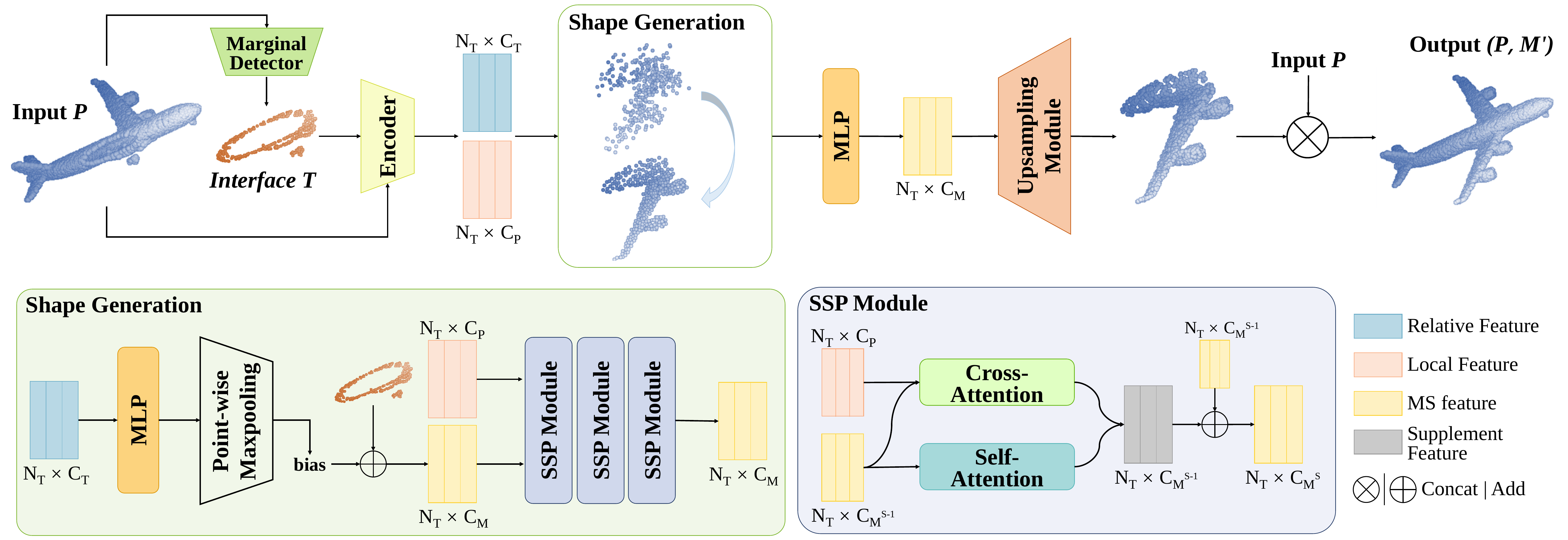}
    \caption{\footnotesize The overall framework of SPAC-Net. Given an input $P$, MAD module is used to localize the \textit{interface} and the Encoder is applied to extract the feature according to $P$. During shape prediction, a coarse shape is generated by learning the displacement from the interface to the missing part. To further facilitate the recovery of details, we leverage several stacked SSP modules to augment structure details before shape upsampling. Finally, we restore missing parts at high resolution by an upsampling module, and add it with input $P$ to obtain the complete point cloud. In the figure, $N_T$ is the number of \textit{interface}, $C_P$, $C_M$ and $C_T$ are the dimension of corresponding feature, respectively.} 
    \label{fig: framework}
\end{figure*}

\subsection{Priors for point cloud completion}
Shape priors have proven to be valuable in various visual tasks \cite{Shape_Prior_1, Shape_Prior_2, Shape_Prior_3, Shape_Prior_4, Shape_Prior_5, Wang_Ang_Lee_2020, PatchComplete}. Wang et al. \cite{Wang_Ang_Lee_2020} propose a feature alignment strategy within a coarse-to-fine pipeline. Their method simultaneously learns to shape priors from both complete and incomplete 3D point cloud, allowing for improved restoration of shape details.
PatchComplete \cite{PatchComplete} learns effective shape priors based on multi-resolution local patches. This approach not only avoids the need to learn 3D shape priors that may be constrained to specific trained categories but also enables geometric reasoning about unseen categories by leveraging local patches.
Zhao et al. \cite{ZhaoZWHK22} present a point cloud completion method conditioned on semantic and spatial relations between different objects. Thus, the other related object servers as a prior, and is treated as an additional input to their model to complete the completion task. 
Zhang et al. \cite{ZhangZWH22} propose an effective completion method that can reduce the solution space based on constrained point movements along light rays. In their approach, the shadow volume generated by the light rays and the missing object can be regarded as a prior, to guide the shape completion. Although these prior-based completion methods have achieved good results, we note that, firstly, this prior information is neither explicit nor clear, and secondly, during the synthesis of rough shapes, they are easily affected by local feature interference.

In this work, we propose a structural prior: \textit{interface} that advocates shifting from the prediction of coarse shape to the estimation of offsets from the interface to the regions of absence. This paradigm not only endows the missing areas with an explicit structural perception preemptively but also circumvents the loss of local feature that arises from deriving coarse shape through global feature. Furthermore, we also investigate the interjection of an auxiliary SSP module between the coarse shape prediction phase and the subsequent shape upsampling process. This module continually refines and optimizes the shape detail information at low resolution, supporting the upsampling module with more accurately detailed shape information.

\section{Method}
\subsection{Problem statement.} 
A completed point cloud contains $N$ points is divided into partial scan $P=\{p_i\}$ and missing part $M=\{m_i\}$, where $p_i$ and $m_i$ denote the ($x$, $y$, $z$) coordinates of the point in their respective regions. Meanwhile,  $P \cap M$ = \{$t_i$\}, $t_i$ represents the coordinates of a  point included in the \textit{interface} $T$. Our goal, from the perspective of completion as missing parts prediction, is to develop a model that can accurately predict a region $M^{\prime}$ that closely resembles the original $M$. The overall architecture of the model is illustrated in Figure~ \ref{fig: framework}.

\begin{figure}[!t]
    \centering
  \setlength{\abovecaptionskip}{1mm} 
    \includegraphics[width = 1\linewidth]{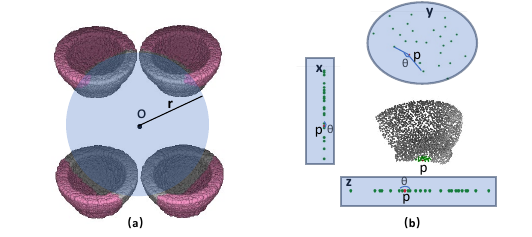}
    \caption{\footnotesize Interface Localization Methods Illustration. (a) and (b) depict the methods under the scenarios of occluded points available and occluded points unknown, respectively.}
    \label{fig:interface_localize}
    \vspace{-10pt}
\end{figure}

\subsection{Interface Localization and Feature Encoding.}
\textbf{Localize.} 
Our aim in this section is to propose a structural prior called \textit{interface} to comprehend the spatial perception of missing parts. We investigate the MAD module to realize the \emph{interface} localizing. 
Due to the existence of different forms of missing data, there are scenarios where occlusion data is available and where occlusion data is unknown. The former can more accurately verify the performance of our method, while the latter can better validate the generalization capability of our method. In the following, we discuss these two situations separately:

1) Occlusion point available: Partial scans in ShapeNet55/34 are obtained by cutting the complete shape of the object with a sphere centered at the occlusion point $\mathbf{o}$ and radius $\mathbf{r}$, which depicts in the Figure \ref{fig:interface_localize}. It is evident that the interfaces are located along the boundary shared by the two. Additionally, points within the interface are closer to the viewpoint than others. Thus, we can calculate the distances between the occlusion point and all input points. By selecting the nearest $N_T$ points in terms of distance from the occlusion point, we can obtain the \emph{interface}. As depicted in the first column of Figure \ref{fig:shapenet55_vis} and Figure \ref{fig:shapenet21_vis}, it is clear that localizing the \emph{interface} in this way is remarkably accurate.

2) Occlusion point unknown: 
The KITTI dataset \cite{KITTI} includes a large number of real-world scenarios, which exhibit more pronounced absences and sparser distributions than those generations in synthetic datasets. In this case, the occlusion point is unknown. 
By definition, the interface lies within the edges of the input, we can devise an edge detection algorithm governed by two threshold values: $\mathbf{r}$ and $\mathbf{\delta}$. Here, $\mathbf{r}$ represents the radius of a sphere centered at $\mathbf{p}$, controlling the number of neighboring points around the central point. Alternatively, $\mathbf{\theta}$ represents the cosine value of the angle formed by any two neighboring points and the central point, projected onto the ($\mathbf{x}$,$\mathbf{y}$,$\mathbf{z}$)  planes. And $\mathbf{\delta}$ determines the maximum permissible area that point $\mathbf{p}$ can enclose when its neighboring points are projected onto the planes. Consequently, we identify points that satisfy the condition where the angle between the two lines does not exceed $\delta$ for both the projections of all neighboring points within radius $r$ onto the ($\mathbf{x}$,$\mathbf{y}$,$\mathbf{z}$)  planes, marking them as part of the interface. To calculate the cosine value of the angle formed between two edges, we use Equation \ref{fm:1}, where $\mathbf{u}$ and $\mathbf{v}$ represent the edge vectors generated by a point and its two different neighboring points. The dot product ($\cdot$) and vector magnitudes ($|\mathbf{u}|$ and $|\mathbf{v}|$) are involved in the computation. Furthermore, we utilize Equation \ref{fm:2} to calculate the distances between each point in set $P$ and other points. The $\mathbf{\delta}$ is set to 0.5 as hyper-parameter. As evident from the first column of Figure \ref{fig:PCN_vis} and Figure \ref{kitti_vis}, the interface extraction is also relatively accurate.

\begin{align}
    \begin{split}\label{fm:1}
        \cos(\theta) &= \frac{{\mathbf{u} \cdot \mathbf{v}}}{{\|\mathbf{u}\| \|\mathbf{v}\|}},
    \end{split}
    \\
    \begin{split}\label{fm:2}
        d(\mathbf{p}, \mathbf{q}) &= \|\mathbf{p} - \mathbf{q}\|.
    \end{split}
\end{align}

\textbf{Encoding.} In the encoder, we incorporate point transformer\cite{Point_Transformer} and set abstraction\cite{CharlesRQi2017PointNetDH} layers to extract local feature $\mathbf{F_P}$ from partial scans, following the approach used in \cite{HaoranZhou2022SeedFormerPS}. Additionally, we utilize EdgeConv \cite{YueWang2018DynamicGC} to capture the relative feature $F_{PT}$ between the partial scans and \textit{interfaces}.

\begin{table*}[t]
	\centering
	\footnotesize
	\setlength{\tabcolsep}{1.91pt}
 \setlength{\belowcaptionskip}{-2mm}
 	\caption{\footnotesize Quantitative results on ShapeNet-55 dataset in terms of L2 Chamfer Distance $\times 1000$ (lower is better) and F-Score@1$\%$.}
	\begin{tabular}{c|cccccccc|ccc|cc}
		\hline\thickhline
		Methods & Table & Chair & Plane & Car & Sofa &Bag &\tabincell{c}{Re\\mote} &\tabincell{c}{Key\\board}
 & CD-S & CD-M & CD-H & CD-Avg & F1  \\
		\hline
		FoldingNet \cite{YaoqingYang2017FoldingNetPC}   	& 2.53 & 2.81 & 1.43 & 1.98 & 2.48 & 2.79 & 1.44 & 1.24 & 2.67 & 2.66 & 4.05 & 3.12 & 0.082 \\
            PCN \cite{WentaoYuan2018PCNPC}  			& 2.13 & 2.29 & 1.02 & 1.85 & 2.06 & 2.86 & 1.33 & 0.89 & 1.94 & 1.96 & 4.08 & 2.66 & 0.133 \\
            TopNet \cite{LynePTchapmi2019TopNetSP}   		& 2.21 & 2.53 & 1.14 & 2.18 & 2.36 & 2.93 & 1.49 & 0.95 & 2.26 & 2.16 & 4.3 & 2.91 & 0.126  \\
            PFNet \cite{ZitianHuang2020PFNetPF}   		& 3.95 & 4.24 & 1.81 & 2.53 & 3.34 & 4.96 & 2.91 & 1.29 & 3.83 & 3.87 & 7.97 & 5.22 & 0.339 \\ 
            GRNet \cite{YingLi2020GRNetGR}   		& 1.63 & 1.88 & 1.02 & 1.64 & 1.72 & 2.06 & 1.09 & 0.89 & 1.35 & 1.71 & 2.85 & 1.97 & 0.238 \\
            SeedFormer \cite{HaoranZhou2022SeedFormerPS}      & 0.72 & 0.81 & 0.40 & 0.89 & 0.71 & 0.79 & 0.36 & 0.45 & 0.50 & 0.77 & 1.49 & 0.92 & 0.472 \\
            PoinTr \cite{XuminYu2021PoinTrDP}   	    & 0.81 & 0.95 & 0.44 & 0.91 & 0.79 & 0.93 & 0.53 & 0.38 & 0.58 & 0.88 & 1.79 & 1.09 & 0.464 \\
		\hline
            PoinTr + \emph{interface}  	    & 0.73 & 0.85 & 0.43 & 0.87 & 0.70 & 0.79 & 0.45 & 0.35 & 0.49 & 0.78 & 1.64 & 0.97 & 0.506 \\
		SPAC-Net & \textbf{0.70} & \textbf{0.81} & \textbf{0.40} & \textbf{0.78} & \textbf{0.66} & \textbf{0.77} & \textbf{0.31} & \textbf{0.43} & \textbf{0.46} & \textbf{0.74} & \textbf{1.43} & \textbf{0.88} & \textbf{0.533} \\
	\hline\thickhline
	\end{tabular}
	\label{table:shapenet55}
\end{table*}

\begin{table*}[ht]
\centering
\caption{\footnotesize Quantitative comparison of SPAC-Net and state-of-the-art methods on ShapeNet-34 in terms of L2 Chamfer Distance $\times 1000$ and F-Score@1$\%$.
(higher is better)} 
  \setlength{\belowcaptionskip}{-2mm}
\setlength{\tabcolsep}{4pt}{
\begin{tabular}{cccccc|ccccc}
\hline\thickhline
\multirow{2}{*}{Methods} &\multicolumn{5}{c}{\textbf{34 seen categories}} & \multicolumn{5}{c}{\textbf{21 unseen categories}} \\
\cmidrule(lr){2-6}\cmidrule(lr){7-11} 
& CD-S & CD-M & CD-H & CD-Avg & F1  & CD-S & CD-M & CD-H & CD-Avg & F1 \\
\hline
FoldingNet~\cite{YaoqingYang2017FoldingNetPC}& 1.86 & 1.81 & 3.38  & 2.35 & 0.139 & 2.76 & 2.74 & 5.36  & 3.62 & 0.095\\
PCN~\cite{WentaoYuan2018PCNPC}               & 1.87 & 1.81 & 2.97  & 2.22 & 0.154 & 3.17 & 3.08 & 5.29  & 3.85 & 0.101\\
TopNet~\cite{LynePTchapmi2019TopNetSP}       & 1.77 & 1.61 & 3.54  & 2.31 & 0.171 & 2.62 & 2.43 & 5.44  & 3.50 & 0.121\\
PFNet~\cite{ZitianHuang2020PFNetPF}          & 3.16 & 3.19 & 7.71  & 4.68 & 0.347 & 5.29 & 5.87 & 13.33 & 8.16 & 0.322\\
GRNet~\cite{YingLi2020GRNetGR}               & 1.26 & 1.39 & 2.57  & 1.74 & 0.251 & 1.85 & 2.25 & 4.87  & 2.99 & 0.216\\
SeedFormer~\cite{HaoranZhou2022SeedFormerPS} & 0.48 & 0.70 & 1.30  & 0.83 & 0.452 & 0.61 & 1.07 & \textbf{2.35}  & \textbf{1.34} & 0.402\\
PoinTr~\cite{XuminYu2021PoinTrDP}            & 0.76 & 1.05 & 1.88  & 1.23 & 0.421 & 1.04 & 1.67 & 3.44  & 2.05 & 0.384\\
\hline
PoinTr + \emph{interface}           & 0.52 & 0.78 & 1.59  & 0.94 & 0.481 & 0.68 & 1.21 & 2.84  & 1.58 & 0.436\\
 SPAC-Net    &  \textbf{0.45}  & \textbf{0.68}  & \textbf{1.24}  & \textbf{0.79} &\textbf{0.548}&\textbf{0.57}&\textbf{1.04}&2.47&1.36&\textbf{0.521}\\
\hline\thickhline
\end{tabular}}
\label{table:ShapeNet-34}
\end{table*}

\subsection{Shape Generation}\label{Section3.3}

\textbf{Coarse generation. } Most existing approaches \cite{WentaoYuan2018PCNPC,YingLi2020GRNetGR} often rely on applying max-pooling on local feature $\mathbf{F_P}$  to predict the coarse shape. However, these approaches, as indicated by Equation \ref{formular: coarse_previous}, can result in the irreversible loss of local feature in subsequent stages. To overcome this, we propose a novel equation based on the \textit{interface} information. Instead of directly reflecting the global feature to the coarse shape, we redefine the coarse shape generation as a movement of \textit{interface} points towards the missing parts. By leveraging the relative positional relationship between the \textit{interface} and the partial scans $\mathbf{F_{PT}}$, we learn the displacement of each \textit{interface} point to its corresponding position in the missing parts. This reformulation is expressed by Equation \ref{formular: coarse_ours}.

{\setlength\abovedisplayskip{5pt}
 \setlength\belowdisplayskip{5pt}
\begin{align}
    \begin{split}
        \label{formular: coarse_previous}
        o_i = {\gamma}({\beta}({\alpha}(F_{P_i}))).
    \end{split}
    \vspace{10pt}
    \\
    \begin{split}
        \label{formular: coarse_ours}
         o_i = {\gamma}({\beta}({\alpha}(F_{P_i T_i}))) +  t_i,
    \end{split}
    \vspace{-10pt}
\end{align}}

where $\mathbf{o_i}$ denotes the coordinates of a point in coarse shape $O$, $\alpha$ represents an MLP maps feature to high dimensions, $\beta$ denotes max-pooling operation and $\gamma$ corresponds to reshape operation.

\textbf{Details augmentation. }Another important consideration is the difficulty of restoring neglected details when performing shape upsampling directly on the coarse shape obtained through Equation \ref{formular: coarse_previous} or Equation \ref{formular: coarse_ours}. To address this issue, we introduce the SSP module, which aims to enhance the augmentation of shape details before shape upsampling. As depicted in the bottom part of Figure~ \ref{fig: framework}, we dynamically update the structural feature of the coarse shape in a multi-stage manner, and each stage is equipped with an SSP module. The initial feature of the missing parts, denoted as $\mathbf{F_{M^0}}$, is obtained by applying an MLP to the concatenation of $O$ (coarse shape) and $\mathbf{F_P}$ (local feature from partial scans). In each stage $s$, the SSP module takes $\mathbf{F_{M^{s-1}}}$ and $\mathbf{F_P}$ as inputs. Then it utilizes self-attention mechanism to capture structure details that require more attention from $\mathbf{F_{M^{s-1}}}$, as well as a cross-attention mechanism to incorporate the details from $\mathbf{F_P}$ that were previously present in the partial scans but might have been overlooked. These newly added structural details are then used as supplemental feature to update the predicted shape and its corresponding feature in the last stage. The process can be summarized as:

{\setlength\abovedisplayskip{0pt}
 \setlength\belowdisplayskip{5pt}
\begin{equation}
    \begin{split}\label{formula: attention}
        h_i = \text{Att}(Q_i W^Q, K_i W^K, V_i W^V), \\
        \text{Uh}(Q, K, V) = {concat}({{\{h}_{i}\}})W^O.
    \end{split}
\end{equation}}

\begin{equation}
    \begin{split}\label{formula: detail augmentation}
        &o^s = \alpha_2(F_{M^s}),\\
        &F_{M^s} = F_{M^{s-1}} + \text{Uh}(Q_1, K_1, V_1)  + \alpha_1(\text{Uh}(Q_2, K_2, V_2)),
    \end{split}
\end{equation}
where Uh is the abbreviation for multi-head attention mechanism, $\alpha_1$ and $\alpha_2$ denote MLP functions, $\mathbf{Q_1}$, $\mathbf{K_1}$ and $\mathbf{V_1}$ obtained by $\mathbf{F_{M^{s-1}}}$, while $\mathbf{Q_2}$ generated by $\mathbf{F_{M^{s-1}}}$, $\mathbf{K_2}$ and $\mathbf{V_2}$ generated by $\mathbf{F_P}$.

It is noteworthy that the MAD module and the SSP module operate as independent entities. The MAD module is tailored for the extraction of interfaces, which are then utilized as structural priors. Our method then transforms the generation of a coarse shape into a displacement from the interface to the missing parts, effectively mitigating the loss of fine-grained details. Conversely, the SSP module builds upon the coarse shape obtained from the interface movement to further refine the precise positioning of each point, yielding a more accurate shape representation at lower resolutions. Despite the divergent design focuses of these modules, their common purpose converges on enhancing the local detail fidelity.

\subsection{Shape Upsampling}
We opt for a simple yet efficient approach \cite{YaoqingYang2017FoldingNetPC}, namely $\sigma$, to perform this process. Some previous methods \cite{WentaoYuan2018PCNPC,YingLi2020GRNetGR} have dedicated upsampling modules designed for shape refinement. However, these modules often face the challenge of simultaneously performing feature upsampling and feature completion. As a result, their performance in both tasks tends to be suboptimal. In contrast to these methods, in Section~\ref{Section3.3}, we have obtained a shape with more structural details. This allows us to alleviate the requirement for the upsampling module to handle feature completion. The equation of the process can be described as follows:
\begin{equation}
    {M^\prime_i} = {\sigma}(F_{M_i}) + o_i.
\end{equation}

\begin{table*}[!ht]
	\centering
	\caption{\footnotesize Quantitative results on PCN benchmark in terms of per-point L1 Chamfer Distance $\times 1000$ (lower is better).}
  \setlength{\belowcaptionskip}{-2mm}
	\footnotesize
	\setlength{\tabcolsep}{3.0pt}
	\begin{tabular}{c|c|cccccccc}
\hline\thickhline
		Methods & Average & Plane & Cabinet & Car & Chair & Lamp & Couch & Table & Boat  \\
		\hline
		FoldingNet \cite{YaoqingYang2017FoldingNetPC}        & 14.31 & 9.49 & 15.80 & 12.61 & 15.55 & 16.41 & 15.97 & 13.65 & 14.99 \\
		TopNet \cite{LynePTchapmi2019TopNetSP}               & 12.15 & 7.61 & 13.31 & 10.90 & 13.82 & 14.44 & 14.78 & 11.22 & 11.12 \\
		AtlasNet \cite{ThibaultGroueix2018AtlasNetAP}        & 10.85 & 6.37 & 11.94 & 10.10 & 12.06 & 12.37 & 12.99 & 10.33 & 10.61 \\
		PCN \cite{WentaoYuan2018PCNPC}                       & 9.64 & 5.50 & 22.70 & 10.63 & 8.70 & 11.00 & 11.34 & 11.68 & 8.59 \\
		GRNet \cite{YingLi2020GRNetGR}                       & 8.83 & 6.45 & 10.37 & 9.45 & 9.41 & 7.96 & 10.51 & 8.44 & 8.04 \\
		CRN \cite{XiaogangWang2020CascadedRN}                & 8.51 & 4.79 & 9.97 & 8.31 & 9.49 & 8.94 & 10.69 & 7.81 & 8.05 \\
		NSFA \cite{WenxiaoZhang2020DetailPP}                 & 8.06 & 4.76 & 10.18 & 8.63 & 8.53 & 7.03 & 10.53 & 7.35 & 7.48 \\
		PMP-Net \cite{XinWen2020PMPNetPC}                    & 8.73 & 5.65 & 11.24 & 9.64 & 9.51 & 6.95 & 10.83 & 8.72 & 7.25 \\
		SnowflakeNet \cite{PengXiang2021SnowflakeNetPC}      & 7.21 & 4.29 & 9.16 & 8.08 & 7.89 & 6.07 & 9.23 & 6.55 & 6.40 \\
  		PoinTr \cite{XuminYu2021PoinTrDP}                    & 8.38 & 4.75 & 10.47 & 8.68 & 9.39 & 7.75 & 10.93 & 7.78 & 7.29 \\
  		SeedFormer \cite{HaoranZhou2022SeedFormerPS}          & 6.74 & 3.85 & 9.05 & 8.06 & 7.06 & \textbf{5.21} & 8.85 & 6.05 & \textbf{5.85} \\
		\hline
  		SPAC-Net          & \textbf{6.61} & \textbf{3.72} & \textbf{8.86} & \textbf{7.58} & \textbf{6.87} & 5.65 & \textbf{8.64} & \textbf{5.61} & 5.97 \\
\hline\thickhline
	\end{tabular}
	\label{table:pcn}
\end{table*}

\begin{figure*}[!ht]
    \centering
    \includegraphics[width = .9\linewidth]{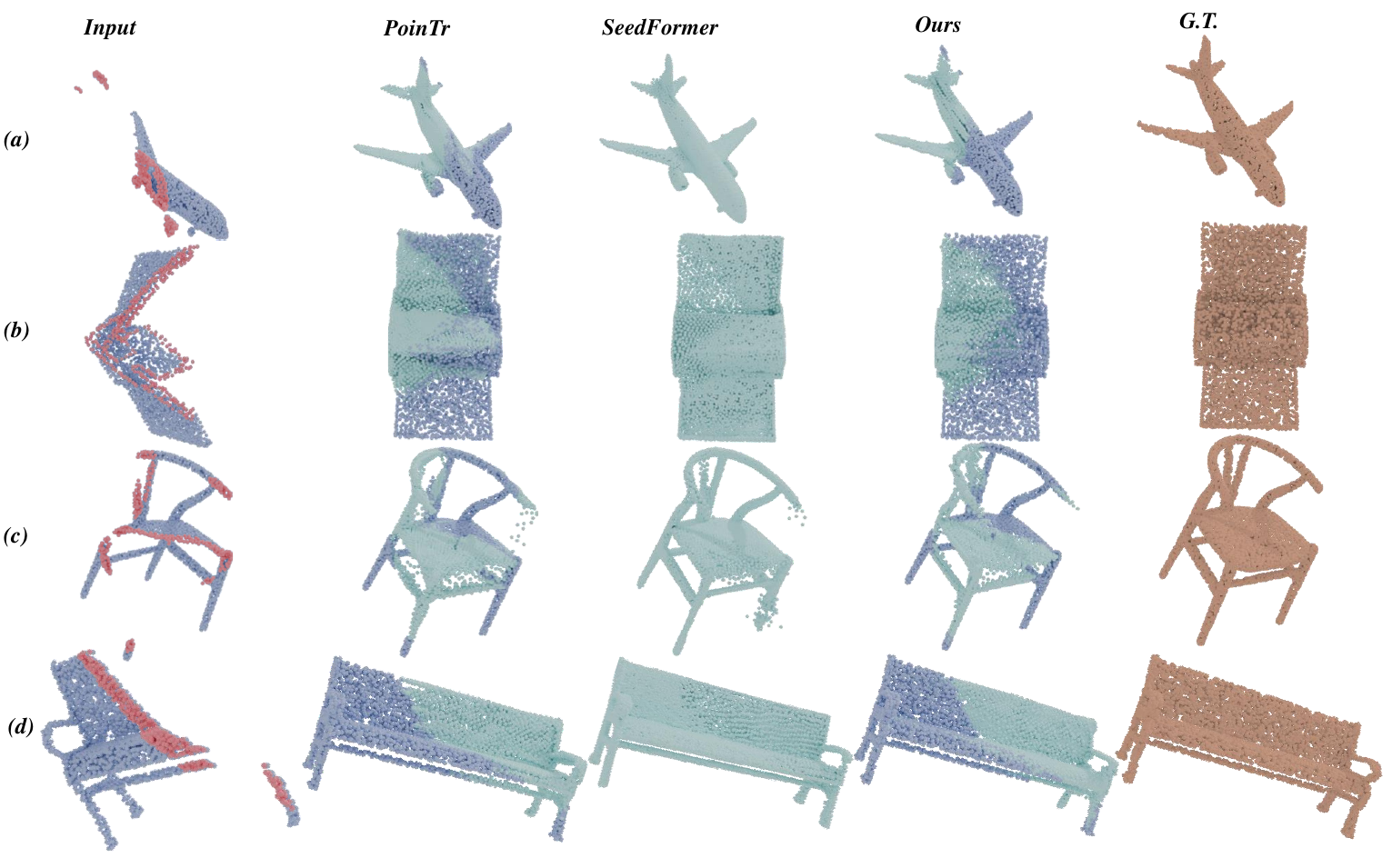}
    \caption{Qualitative comparison of SPAC-Net and state-of-art methods on ShapeNet-55 dataset, showing Airplane, Printer, Chair and Bench from top to bottom.The blue points indicate partial scans, the cyan points signify the predicted results, the magenta points denote the interfaces, and the brown points correspond to the ground truth.}
    \label{fig:shapenet55_vis}
\end{figure*}

\subsection{Joint Loss}
We choose symetric Chamfer Distance (CD) as loss function to supervise our prediction as close as groundtruth. In practice, we design our joint loss $\Gamma$ into the addition of ${\Gamma}_{partial}$ and ${\Gamma}_{complete}$, where ${\Gamma}_{partial}$ denotes the average chamfer distance between coarse prediction and groundtruth, ${\Gamma}_{complete}$ denotes the average distance between final prediction and groundtruth. The function is written as:
\begin{equation}
    {\Gamma} = \lambda_1{\Gamma}_{partial} + \lambda_2{\Gamma}_{complete},
\end{equation}
where $\lambda_1$ and $\lambda_2$ are the hyper-parameters for balancing the importance of corresponding loss function in the training process, and we set $\lambda_1$=$\lambda_2$=1 in our experiments.

\begin{figure*}[t]
    \centering
    \includegraphics[width =1\linewidth]{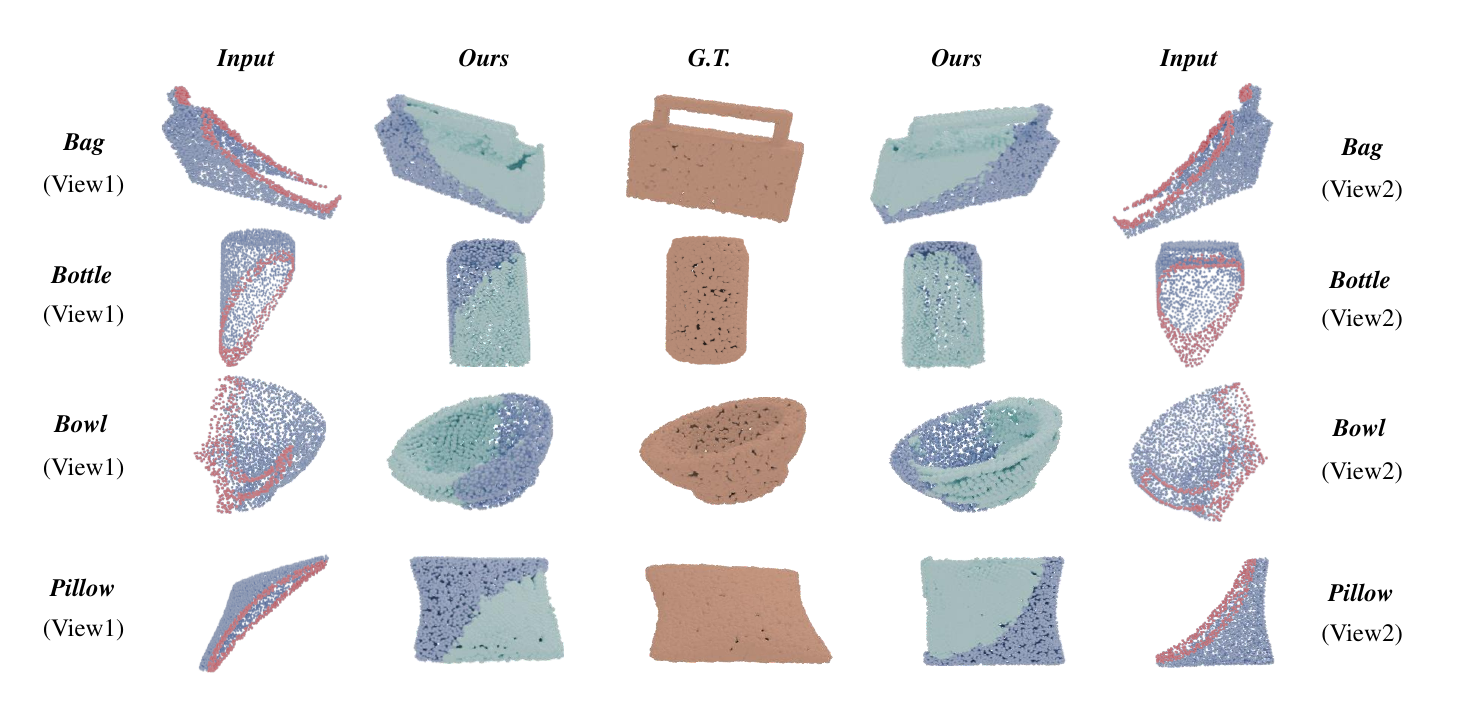}
    \caption{Qualitative results of our method on some novel objects included in ShapeNet-34 dataset. For each object, we provide completion results in two occlusion cases. The blue points indicate partial scans, the cyan points signify the predicted results, the magenta points denote the interfaces, and the brown points correspond to the ground truth.}
    \label{fig:shapenet21_vis}
\end{figure*}

\section{Experiments}
\subsection{Dataset and evaluate metrics.}\label{sec4.1}
\textbf{Dataset.}  
1) ShapeNet-55/34 benchmark \cite{XuminYu2021PoinTrDP}: In the pre-processing step, we divide the groundtruth into an input point cloud containing 2048 points and an output point cloud containing 6144 points, following the approach described in \cite{XuminYu2021PoinTrDP}. To ensure a diverse training dataset, we initiate the process by randomly selecting a viewpoint and then proceed to eliminate the $N$ points that are most distant from this perspective. The remaining points are subsequently downsampled to a uniform count of 2048, yielding a partial point cloud suitable for training purposes. This means that the missing part of each sample may vary in each training epoch. During testing, we use 8 fixed viewpoints and set the count of masked points $N$ to 2048, 4096, or 6144, representing easy, medium, and hard levels (25\%, 50\%, or 75\% of the complete point cloud). 
2) PCN benchmark \cite{WentaoYuan2018PCNPC}: The PCN benchmark consists of 30714 objects from 8 categories. Each complete object shape contains 16384 uniformly sampled points obtained from the CAD model. With the aim of aligning the input distribution more closely with that of real-world sensor data, the partial point clouds are generated by back-projecting 2.5D depth images into 3D, and the partial scans are composed of 2048 points. This benchmark presents a challenging task due to its high-resolution completion requirements. To ensure a fair comparison, we maintain consistent experimental configurations with the approach proposed in \cite{WentaoYuan2018PCNPC}. 
3) KITTI benchmark \cite{KITTI}: The KITTI benchmark features 11 cars within a dataset of 425 real-world frames, identified by 3D bounding boxes. These objects exhibit more pronounced absences and sparser distributions than those generations in synthetic datasets. Completing the point cloud in this benchmark poses a challenging task, requiring robust methods to handle the inherent complexities.

\begin{table*}[!ht]
	\centering
 	\caption{\footnotesize Completion results on KITTI  dataset evaluated as Fidelity and MMD. Lower is better.}

	\footnotesize
	\setlength{\tabcolsep}{4mm}
        \vspace{5pt}
	\begin{tabular}{c|cccccc|c}
\hline\thickhline
		 & PCN  
          & FoldingNet  
          & TopNet  
          & GRNet 
          & PoinTr 
          & SeedFormer 
          & Ours\\
		\hline
        Fidelity  & 2.235 & 7.467 & 5.354  & 0.816  & 0.000 & 0.151 & \textbf{0.000}\\
        MMD       & 1.366 & 0.537 & 0.636  & 0.568  & 0.526 & 0.516 & \textbf{0.473} \\
\hline\thickhline
	\end{tabular}
	\label{table:kitti}
\end{table*}

\begin{figure*}[t]
    \centering
       \setlength{\belowcaptionskip}{-4mm}
    \includegraphics[width = 1.0\linewidth]{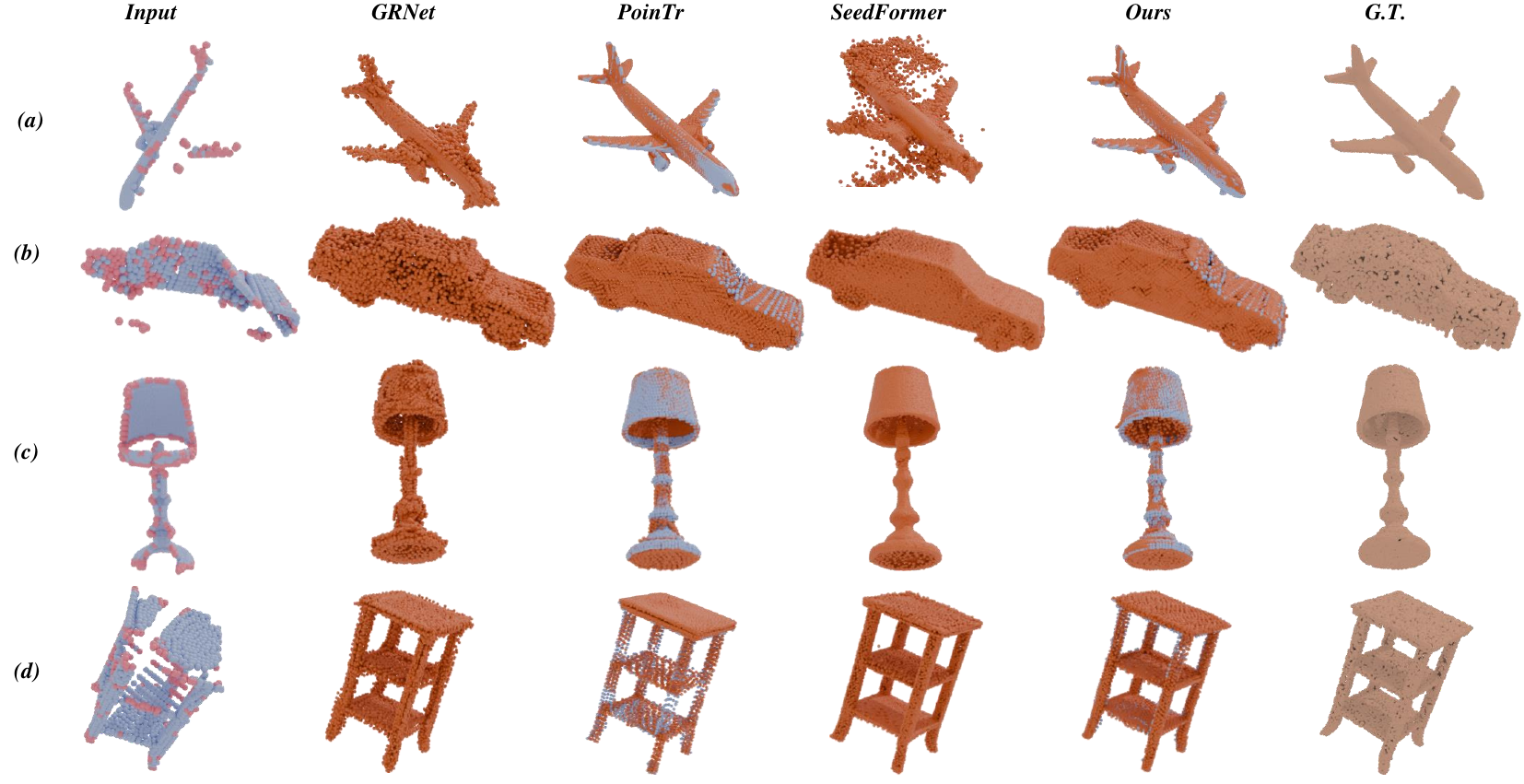}
    \caption{
    The qualitative comparison of our method with other SOTA methods on the PCN dataset \cite{WentaoYuan2018PCNPC}. The leftmost column shows the input data, and the rightmost column displays the ground truth. The middle columns show the visual results of our method and other methods. As can be seen, our method produces results that are closer to the ground truth.}

    \label{fig:PCN_vis}
\end{figure*}

\textbf{Metrics.}
1) Chamfer distance: We employ two versions of the Chamfer distance, CD-$\ell_1$ and CD-$\ell_2$, as evaluation metrics to measure the similarity between two objects in terms of their overall shape. These metrics are widely used in previous works and provide a quantitative measure for comparing the performance of our model with existing methods.
2) F-Score: Following the approach proposed in \cite{f-score}, we utilize the F-Score metric to assess the accuracy of our reconstructed point cloud. 
3) Fidelity: It represents the mean distance from each input point to its nearest counterpart in the output, offering insights into the fidelity of our completion results in maintaining the integrity of the original input shape.

4) Minimal Matching Distance (MMD): The MMD is calculated using the Chamfer Distance-$\ell_2$ (CD-$\ell_2$) between the generated point cloud and the nearest car point cloud from the ShapeNet \cite{ShapeNet} dataset. This metric evaluates the degree to which the completed shape aligns with the canonical form of a typical car shape.

\subsection{Implementation details.}\label{sec4.2}
\textbf{Training Details.} 
Our model is implemented using the PyTorch library, and all experiments are performed on a single Nvidia RTX 3090 GPU. We utilize the ADAMW\cite{ADAMW} optimizer to train our model, with an initial learning rate of 0.0005. We also apply a decay rate of 0.0005 for every epoch to optimize the learning process.
To handle the complexity of localizing the \textit{interface}, we conduct pre-processing on the input data in order to reduce the computational time during both the training and testing phases.

\textbf{Data pre-processing and hyper-parameters.}  
For feature encoding, the input size is 2048 to both the Shapenet-55/34 dataset and PCN dataset \cite{WentaoYuan2018PCNPC}. Following the SeedFormer \cite{HaoranZhou2022SeedFormerPS}, the input undergoes a series of transformations: ($N_P$=2048, $C_P$=3) $\longrightarrow$ ($N_P$=512, $C_P$=128) $\longrightarrow$ ($N_P$=$N_T$, $C_P$=256), resulting in aggregated local feature $F_P$ for $N_P$ downsampled patches from partial scans. Next, we employ the edgeconv \cite{YueWang2018DynamicGC} module to extract relative feature between interfaces and partial scans, following the specific procedure outlined as follows: edgeconv($N$=$N_T$, $C_T$=3)$ \longrightarrow$  edgeconv($N$=$N_T$, $C_T$=64)$ \longrightarrow $edgeconv($N$=$N_T$, $ C_T$=128) $\longrightarrow$ edgeconv($N$=$N_T$, $C_T$=256).
The grid size of FoldingNet \cite{YaoqingYang2017FoldingNetPC} is set to ($-1.$ , $1.$). For the ShapeNet-55/34 dataset, the upsampling factor $r$ is set to $16$, resulting in an output with $N_{M^\prime}$=6144 points. As for the PCN dataset, the $r$=$64$, yielding an output with $N_{M^\prime}$=14336 points.

\subsection{Evaluation on Synthetic Datasets.}

In this section, to validate the effectiveness of the \emph{interface} localization, we first proceed with the case where the occlusion viewpoint is known. Utilizing the known observed viewpoint during the generation of partial scans on the ShapeNet55/34 dataset \cite{XuminYu2021PoinTrDP}, we aligned with the method illustrated in Fig \ref{fig:interface_localize}(a) to identify the interface. To ensure a fair comparison, we also incorporated the interface into the baseline PoinTr \cite{XuminYu2021PoinTrDP} to verify the efficacy of integrating such prior knowledge. Subsequently, we conducted ablation studies to compare different definitions of the interface. In real-world scenarios, where the intricacies of the environment make the occlusion viewpoint not directly obtainable, we employed the method depicted in Figure \ref{fig:interface_localize}(b) to pinpoint the interface on the PCN\cite{WentaoYuan2018PCNPC} dataset, under generalizing conditions where the occlusion viewpoint is indeterminate. 

\begin{figure*}[!ht]
    \centering
     \setlength{\belowcaptionskip}{-3mm}
    \includegraphics[width = 1.0\linewidth]{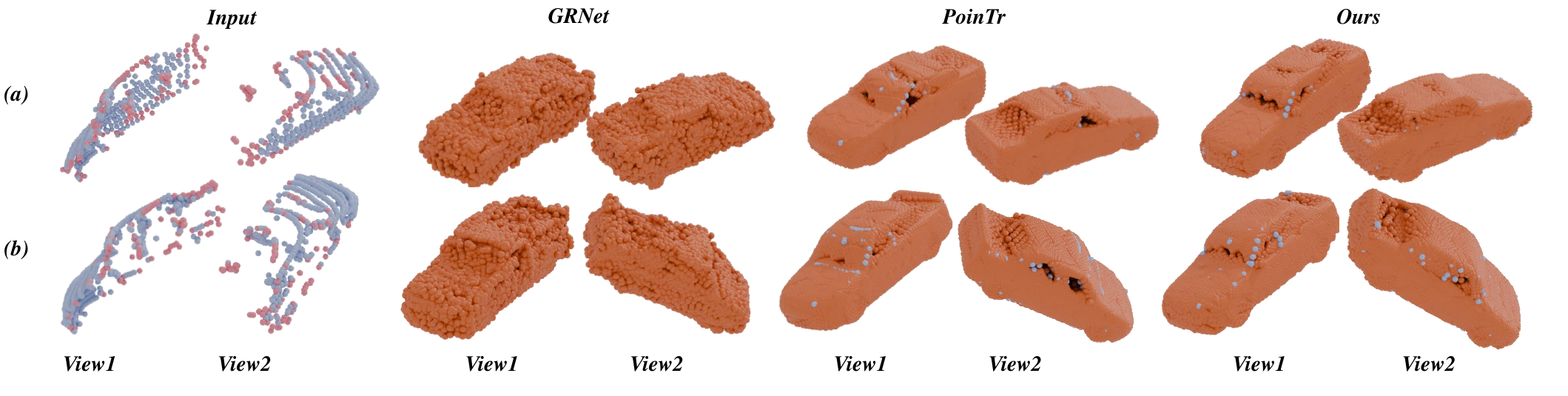}
    \caption{
    The qualitative comparison of our method with other SOTA methods on the KITTI dataset \cite{KITTI}. To better illustrate the differences in detail between our method and others, we provide two views of the results.}
    \label{kitti_vis}
\end{figure*}

\begin{table}[!t]
  \centering
    \caption{\footnotesize Complexity evaluation on the PCN dataset, focusing on two critical metrics: the count of trainable parameters and the theoretical computational complexity in terms of FLOPs cost.}
  \renewcommand\arraystretch{1.1}
  \setlength{\tabcolsep}{5pt}{
  \begin{tabular}{c|ccccc|c}
\hline\thickhline
   Methods & Params  & FLOPs\\
  \hline
    FoldingNet\cite{YaoqingYang2017FoldingNetPC} & \textbf{2.41M} & 27.65G \\
    PCN\cite{WentaoYuan2018PCNPC} & 6.84M & 14.69G \\
    GRNet\cite{YaoqingYang2017FoldingNetPC} & 76.71M & 25.88G \\
    PoinTr\cite{XuminYu2021PoinTrDP} & 31.27M & 10.59G \\
    SeedFormer\cite{HaoranZhou2022SeedFormerPS} & 3.20M & 29.61G \\
  \hline
    SPAC-Net & 8.65M & \textbf{5.21G} \\
\hline\thickhline
  \end{tabular}}
    \label{table: complexity analysis}
    \end{table}

\begin{table}[!t]
  \centering
    \caption{\footnotesize Ablation study on 
    \textit{interface} hyper-parameter choices.}
  \renewcommand\arraystretch{1.1}

  \setlength{\tabcolsep}{5pt}{
  \begin{tabular}{c|c|c}
\hline\thickhline
   Methods & Attempts  & CD\\
  \hline
  \multirow{2}{*}{w/o \textit{interface}} & \multirow{2}{*}{global feature} & \multirow{2}{*}{1.04} \\
  & & \\
  \hline
  \multirow{2}{*}{w/ \textit{interface}} & intersection & \textbf{0.88}\\
                               & downsampling points & 0.97\\
\hline\thickhline
  \end{tabular}}
    \label{table: ablate interface}
    \end{table}

  \begin{table}[!t]
  \centering
    \caption{\footnotesize Ablation study on the value of $\delta$ during localization interface.}
  \renewcommand\arraystretch{1.1}
  \setlength{\tabcolsep}{4mm}{
  \begin{tabular}{c|c|c}
\hline\thickhline
    Case   & Value    & CD\\
  \hline
   ${\delta}_0$       & 0.3         & 7.03\\
   ${\delta}_1$       & 0.4         & 6.82\\
   ${\delta}_2$       & 0.5         & \textbf{6.61}\\
   ${\delta}_3$       & 0.6         & 6.92\\
   ${\delta}_4$       & 0.7         & 6.85\\
\hline\thickhline
  \end{tabular}}
    \label{table: ablate delta}
\end{table}

  \begin{table}[!t]
  \centering
    \caption{\footnotesize Ablation study on the number of SSP modules}.
  \renewcommand\arraystretch{1.1}

  \setlength{\tabcolsep}{4mm}{
  \begin{tabular}{c|c|c}
\hline\thickhline
   Model   & number    & CD\\
  \hline
   A       & 0         & 1.12\\
   B       & 1         & 1.02\\
   C       & 2         & 0.91\\
   D       & 3         & \textbf{0.88}\\
\hline\thickhline
  \end{tabular}}
    \label{table: ablate SSP}
\end{table}

\textbf{Results on ShapeNet-55 benchmark.} Table \ref{table:shapenet55} presents comparative results of SPAC-Net with a variant of PoinTr utilizing \textit{interface} termed "PoinTr + \textit{interface}", alongside previous state-of-the-art methodologies in terms of average CD-$\ell_2$. The table shows the results for some categories (Table, Chair, Plane, Car, Sofa, Remote, and Keyboard) as well as the overall performance across all 55 categories. From the results, it is evident that both the SPAC-Net and "PoinTr + \textit{interface} outperform PoinTr \cite{XuminYu2021PoinTrDP} in all aspects. Notably, our method achieves better performance than SeedFormer\cite{HaoranZhou2022SeedFormerPS} with an improvement of $4.35\%$ in overall CD-$\ell_2$ and $11.44\%$ in terms of F-Score. These results indicate that objects completed using our method not only closely resemble the ground truth in terms of overall shape but also exhibit superior accuracy in preserving the position of each point.
Therefore, employing the interface as a structural prior to establishing a sense of the geometric structure of the missing regions is effective in guiding the process of point cloud completion.

Figure \ref{fig:shapenet55_vis} showcases the qualitative experimental results and detailed information regarding the quantitative evaluation of the Shapenet-55 dataset. In the figure, we present a comparative visualization between SPAC-Net and existing state-of-the-art methods such as PoinTr \cite{XuminYu2021PoinTrDP} and SeedFormer\cite{HaoranZhou2022SeedFormerPS}, utilizing pre-trained models provided by these methods. The results clearly demonstrate that our approach not only excels in detail completion but also exhibits superior coherence between the missing regions and the input.

\textbf{Results on ShapeNet-34 benchmark.} We further evaluate the performance of the methods on the ShapeNet-34 dataset, which consists of 34 seen categories and 21 unseen categories. This allows us to assess the generalization ability of the methods on novel categories. Table \ref{table:ShapeNet-34} presents the results for two testing cases, measured by CD-$\ell_2$ and F1-Score. Our proposed method demonstrates superior performance compared to previous state-of-the-art methods across almost all metrics and settings. And the variant PoinTr +\emph{ interface} also has exhibited a modest improvement in performance over the PoinTr\cite{XuminYu2021PoinTrDP} baseline. Notably, in terms of F1-Score, our method outperforms SeedFormer \cite{HaoranZhou2022SeedFormerPS}, the best-performing method, by 17.5\% and 22.8\% on the 34 seen categories and 21 unseen categories.

In Figure \ref{fig:shapenet21_vis}, we showcase the qualitative results on ShapeNet-21 dataset, which contains the completion results of our method on several novel categories. To more intuitively reflect the generalization ability of our method, for each object, we provide completion results in two occlusion cases. The performance of these novel objects demonstrates the outstanding generalization capability of our method.

\textbf{Results on PCN benchmark.} The PCN benchmark is widely recognized as a challenging benchmark for point cloud completion at higher resolution. Our method achieves remarkable results, surpassing other methods in 6 out of 8 categories based on the CD-$\ell_1$ metric, as demonstrated in Table \ref{table:pcn}. This indicates that our method excels in restoring objects to higher resolution, bringing them closer to the ground truth. Figure \ref{fig:PCN_vis} showcases the qualitative results, highlighting the effectiveness of our method in capturing sensitive details, similar to other methods. Moreover, our method demonstrates the ability to capture sensitive details that other methods may neglect and the robustness in completing objects across different categories. Overall, our method achieves superior performance in terms of restoring objects at higher resolution, capturing fine details, and ensuring stability across different object categories.

\subsection{Evaluation on Real-World Benchmark.}
\textbf{Results on KITTI benchmark. } When evaluating the absence of ground truth references in the KITTI benchmark, we first perform fine-tuning of our model on the Cars of PCN dataset\cite{WentaoYuan2018PCNPC} following the experimental settings of \cite{YingLi2020GRNetGR}. We then utilize Fidelity and MMD as evaluation metrics to assess the quality of our predictions. According to the quantitative comparison in Table \ref{table:kitti}, both our method and PoinTr \cite{XuminYu2021PoinTrDP} achieve the Fidelity of 0s. This suggests that our method focuses solely on predicting the missing parts. Moreover, by comparing the MMD values and examining the qualitative visualizations in Figure \ref{kitti_vis}, it can be observed that our method excels at accurately reconstructing the complete shape and details of a typical car, even in scenarios involving severe occlusion in partial scans.

\subsection{Complexity Analysis.}
We conduct a comprehensive complexity analysis of our method. In Table 5, we benchmarked our method against other state-of-the-art methods using the PCN dataset, focusing on training parameter consumption and FLOPs (Floating Point Operations per Second). The results show that our method outperforms the others in terms of FLOPs.

\subsection{Ablation Studies.} 
In this section, to verify the effectiveness of our key modules, we conduct ablation studies on the proposed MAD and SSP modules.

\textbf{MAD module.} In order to examine the effectiveness of our MAD module, we replace the generation process with two alternative strategies on ShapeNet55 benchmark. Firstly, we replace the procedure of generating the coarse shape of our method with the global feature obtained via max-pooling. Secondly, we redefine the \textit{interface} as a set of points obtained by downsampling on partial scans, instead of the intersection between partial scans and missing parts. The quantitative experimental results, as shown in Table \ref{table: ablate interface}, indicate that generating the coarse shape through Equation \ref{formular: coarse_ours} yields better results compared to generating it through Equation \ref{formular: coarse_previous}. Furthermore, it is optimal to define points in the interface as the intersection between the two regions.

\textbf{The hyper-parameter \textbf{$\delta$.}} 
Due to the absence of ground truth values for \emph{interface} in the PCN dataset, a direct evaluation of the accuracy of the edge detection algorithm within the MAD module is not possible. Therefore, we assessed the performance enhancement of the interface on the CD-$\ell_1$ metric using four distinct $\delta$ values on the PCN dataset. The results, presented in Table \ref{table: ablate delta}, indicate that the optimal performance of CD-$\ell_1$ for SPAC-Net on~the~PCN~dataset~is~achieved~when~the~$\delta$~value is set to 0.5.

\textbf{Structure Supplement Module. } To evaluate the importance and optimal quantity of the SSP modules in our network, we also conduct an ablation study by removing the specific SSP modules from our framework. Table \ref{table: ablate SSP} showcases the results. From the table, when one SSP module is removed from the network, there is a direct decrease in network performance. Furthermore, the results indicate that the network achieves the best performance when three SSP modules are used.

\section{Conclusion and Discussions}
In this paper, we introduce the concept of \textit{interface} as a structural prior to enhancing the perception of missing parts in point cloud completion. Guided by this structural prior, we designed a novel network architecture called SPAC-Net for the shape completion task. Extensive experiments demonstrate that our method outperforms state-of-the-art approaches on benchmark datasets. Looking ahead, we believe that the concept of structural prior can be applied to other point cloud downstream tasks, such as self-supervised point cloud understanding, point cloud denoising, and more.

\begin{figure*}[!ht]
    \centering
     \setlength{\belowcaptionskip}{-3mm}
    \includegraphics[width = 1.0\linewidth]{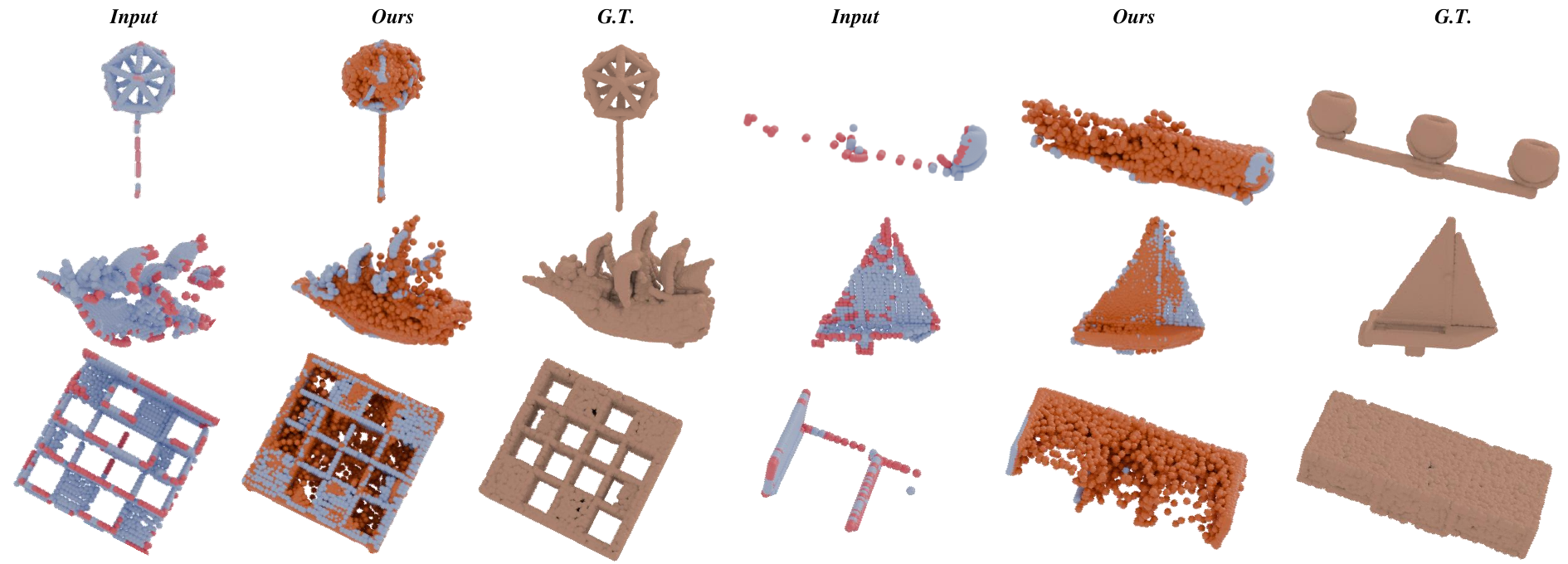}
    \caption{The visualization of some failure cases. Our method struggles to restore the ideal shape if the input data is severely occluded. }
    \label{bad_case}
\end{figure*}

Regarding the limitations of our method, if the given training samples are limited or if the test objects are severely occluded, our current approach may struggle to restore the original shape effectively. Figure~\ref{bad_case} illustrates some examples. Limited training samples can impair the feature learning's perception of object structure and geometric details, while severe occlusion can introduce instability and noise in predicting the completion region. Nonetheless, these challenges are common in the field. Additionally, we have experimented with automatic interface localization using an end-to-end architecture for the completion task. However, our results indicated that the automatic interface localization did not perform as well as the MAD method proposed in this paper. These limitations highlight potential directions for future research.


\bibliographystyle{abbrv-doi}

\bibliography{template}

\end{document}